%% file: main.tex
\def\year{2020}\relax
\documentclass[letterpaper]{article} 

\usepackage{aaai20}  
\usepackage{times}  
\usepackage{helvet} 
\usepackage{courier}  
\usepackage[hyphens]{url}  
\usepackage{graphicx} 
\usepackage{graphicx}  
\nocopyright

\usepackage{amsmath}
\usepackage{mathtools}
\usepackage{amsfonts}

\usepackage{float}

\usepackage{caption}
\usepackage{subcaption}

\usepackage{array}

\newcommand{\mypar}[1]{\textbf{#1}}

\usepackage[capitalize,noabbrev]{cleveref}

\usepackage{microtype} 

\title{Learning Domain-Independent Planning Heuristics with Hypergraph Networks}

\author{
 William Shen\textsuperscript{\rm 1}, Felipe Trevizan\textsuperscript{\rm 2}, Sylvie Thi\'ebaux\textsuperscript{\rm 2}\\
 \textsuperscript{\rm 1,\rm 2}The Australian National University
 \\
 \textsuperscript{\rm 1}william.w.y.shen@gmail.com\hspace{4pt}
 \textsuperscript{\rm 2}first.last@anu.edu.au
}

\newcommand\MyHGNFull{\textsc{STRIPS-HGNs}}
\newcommand\MyHGN{\textsc{STRIPS-HGN}}

\newcommand\hSpatial{$h^{HGN}$}
\newcommand\hAdd{$h^{add}$}
\newcommand\hMax{$h^{max}$}

\newcommand{\citet}[1]{\citeauthor{#1} \shortcite{#1}}
\newcommand{\citep}{\cite}
\newcommand{\citealp}[1]{\citeauthor{#1} \citeyear{#1}}

\begin{document}

\maketitle

\begin{abstract}
We present the first approach capable of learning domain-independent planning
heuristics entirely from scratch.
The heuristics we learn map the hypergraph representation of the
delete-relaxation of the planning problem at hand, to a cost estimate that
approximates that of the least-cost path from the current state to the goal
through the hypergraph.
We generalise Graph Networks to obtain a new framework for learning over
hypergraphs, which we specialise to learn planning heuristics by training over
state/value pairs obtained from optimal cost plans.
Our experiments show that the resulting architecture, \MyHGNFull{}, is capable
of learning heuristics that are competitive with existing delete-relaxation
heuristics including LM-cut.
We show that the heuristics we learn are able to generalise across different
problems and domains, including to domains that were not seen during training.
\end{abstract}

\input{introduction.tex}

\section{Planning Heuristics}
We are concerned classical planning problems represented in propositional STRIPS
\cite{fikes-nilsson-1971-strips}. Such a problem is a tuple $P = \langle  F, O,
I, G, c \rangle$ where
$F$ is the set of propositions; $O$ is the set of actions; $I \subseteq F$
represents the initial state; $G \subseteq F$ represents the set of goal states;
and $c(o)$ is the cost of action $o \in O$.
Each action $o \in O$ is defined as a triple $\langle Pre(o), Add(o),
Del(o)\rangle$ where the precondition $Pre(o)$ is the set of propositions which
must be true in order for $o$ to be applied, while the add- and delete-effects
$Add(o)$ and $Del(o)$ are the sets of propositions which the action makes true
and false, respectively, when applied.

A solution plan $\pi = o_1, \dots, o_n$ for a STRIPS problem is a sequence of
applicable actions leading from the initial state to the goal, i.e., $\pi$
induces a sequence of states $s_1, \ldots, s_{n+1}$ such that $s_1\!=\!I$,
$G\subseteq s_{n+1}$, and for all $i\!\in\! \{1,\ldots,n\}$ $s_{i+1} =
(s_i\setminus Del(o_i)) \cup Add(o_i)$ and $Pre(o_i) \subseteq s_{i}$.
The cost of a plan is the sum of the costs of its actions $\sum_{i \in \{1,
\dots, n\}} c(o_i)$.
An optimal plan is a plan which has minimum cost.

\mypar{Heuristics.}
Let $\mathcal{S} \subseteq 2^{F}$ be the state space. 
A heuristic function $h\colon \mathcal{S} \to \mathbb{R}$ provides an estimate
of the cost to reach a goal state from a state $s$, allowing a search algorithm
to focus on promising parts of the state space.
The optimal heuristic $h^*(s)$ is the heuristic that gives the cost of the
optimal plan to reach a goal state from $s$. 
A heuristic $h$ is admissible iff it never overestimates this optimal cost,
i.e., $\ h(s) \leq h^{*}(s) \ \forall s \in \mathcal{S}$, and is inadmissible
otherwise.
Many heuristics are obtained by approximating the cost of the optimal plan for a
relaxation of the original problem $P$. A well-known relaxation, the {\em
delete-relaxation} $P^+$ of $P$ is obtained by ignoring the delete-effects
$Del(o)$ of all actions in $P$, i.e.,  $P^+ = \langle  F, O', I, G, c \rangle$,
where $O' = \{\langle Pre(o), Add(o), \emptyset \rangle \mid o \in O \}$.
This works considers three baseline domain-independent heuristics which are
based on the delete-relaxation: $h^{max}$ (admissible),  $h^{add}$
(inadmissible) \cite{bonet-geffner-2001-search}, and the Landmark-Cut heuristic
(admissible) \cite{helmert-domshlak-2009-lmcut}.

\input{hypergraph-networks.tex}

\input{strips-hgn.tex}

\section{Empirical Evaluation}
\label{sec:evaluation}
Our experiments are aimed at showing the generalisation capability of
\MyHGNFull{} to problems they were not trained on. 
For each experiment, we select a small pool of training problems (potentially
from several domains) and train a \MyHGN{}. 
We then evaluate the learned heuristic on a larger pool of testing problems with
differing initial/goal states, problem sizes and even domains.
We repeat each experiment for \MyHGNFull{} 10 times, resulting in 10 different
trained networks, to measure the influence of the randomly generated problems
and the training procedure.

\subsection{Experimental Setup}\label{subsec:hgn-conf-exp}

\mypar{Hardware.} All experiments were conducted on an Amazon Web Services \texttt{c5.2xlarge}
server with an Intel Xeon Platinum 8000 series processor running at 3.4Ghz.
To ensure fairness between \MyHGN{} and our baselines, each experiment was
limited to a single core.
We enforced a 16GB memory cutoff; however, only blind search reached this cutoff
and the other planners never exceeded 2GB.

\mypar{Search Configuration.}
We compare \MyHGNFull{} against the following baselines: no heuristic (i.e.,
blind), \hMax{}, LM-cut, and \hAdd{}.
These baselines all represent heuristics computable using the same input as used
by \MyHGNFull --- the delete-relaxation hypergraph --- making this a fair
comparison as all heuristics have access to the same information.
We use A* as the search algorithm to compare the different heuristics, since
\MyHGNFull{} is trained using optimal heuristic values and we believe that its
estimates are sufficiently informative to find the optimal solution.

To generate the training data for each training problem, we used Fast Downward
\cite{helmert-2006-fastdownward} configured with A* search and the LM-cut
heuristic with a timeout of 2 minutes. 
To evaluate each testing problem with a heuristic, we used A* search in
Pyperplan \cite{alkhazraji-etal-2011-pyperplan} with a 5 minute timeout. For
each problem and heuristic, we run A* once.

We used Pyperplan for evaluation as \MyHGNFull{} are implemented in Python. We
observed that the implementations of the delete-relaxation heuristics in
Pyperplan are much slower than their counterparts in Fast Downward. 
Hence, our results for CPU times should be considered as preliminary.

\mypar{\MyHGNFull{} Configuration.}
We generate the hypergraph of each planning problem by using the delete-relaxed
problem computed by Pyperplan. 
For a STRIPS problem $P = \langle  F, O, I, G, c \rangle$ and a given state $s
\subseteq F$, we encode the input features for each proposition (vertex) $p \in
F$ as a vector $[x_s, x_g]$ of length 2 where: $x_s=1$ (resp. $x_g=1$) iff $p$
is true in state $s$ (resp. in the goal $G$), and 0 otherwise.
The input feature for each action $o \in O$ represented by a hyperedge $e$ is a
vector $[w_e, r_e, s_e]$, where $w_e$ is the cost $c(o)$ of $o$, and $r_e =
\vert Add(o) \vert$ and $s_e = \vert Pre(o) \vert$ are the number of positive
effects and preconditions for action $o$, respectively. 
$r_e$ and $s_e$ are used by a \MyHGN{} to determine how much of a `signal' it
should send from a given hyperedge.

We set the number of message passing steps $M$ for the recurrent core HGN block
to 10, and implement each update function as a Multilayer Perceptron (MLP) with
two sequential fully-connected (FC) layers, each with an output dimensionality
of 32. 
We apply the LeakyReLU activation function \cite{maas-etal-2013-leakyrelu}
following each FC layer.
We add an extra FC layer with an output dimensionality of 1 in $\phi^u$ of the
decoding block. Since the input to a MLP must be a fixed-size vector, we
concatenate each update function's input features before feeding them into the
MLP. 

However, for the hyperedge update function $\phi^e$ in the core block, the
number of receiver $r_e$ and sender $s_e$ vertices may vary with each hyperedge.
For a given set of domains, we can compute the maximum number of preconditions
$N_{sender}$ and positive effects $N_{receiver}$ of each possible action by
analysing their action schemas -- this allows us to fix the size of the feature
vectors for the receiver and sender vertices.
We convert the set of input vertices $\mathbf{R}_k$ (resp. $\mathbf{S}_k$) into
a fixed-size vector determined by $N_{receiver}$ (resp. $N_{sender}$), by
stacking each vertex feature $\mathbf{v} \in \mathbf{R}_k$ (resp. $\mathbf{v}
\in \mathbf{S}_k$) in alphabetical order by their proposition names, and padding
the vector with zeros if the required length is not reached.

For the aggregation functions $\rho^{e \rightarrow v}$, $\rho^{e \rightarrow u}$
and $\rho^{v \rightarrow u}$ in the core block of a \MyHGN{}, we use
element-wise summation. We denote the heuristic learned by this configuration of
\MyHGN{} as \hSpatial{}.

\mypar{Training Procedure.}
We split the training data into $n$ bins using quantile binning of the target
heuristic values and use stratified $k$-fold to split the training set into
folds $F = \{f_1, \dots, f_k\}$ with each fold containing approximately the same
percentage of samples for each heuristic bin.
For each fold $f \in F$, we train a \MyHGN{} using $F \setminus f$ as the
training set and $f$ as the validation set, and select the network at the epoch
which achieved the lowest loss on the validation set $f$.
Since we train one \MyHGN{} for each of the $k$ folds, we are left with $k$
separate networks.
We select the network which performed best on its validation set as the single
representative \MyHGN{} for an experiment, which we then evaluate on a
previously unseen test set.

Although $k$-fold is more commonly used for cross validation, we use it to
reduce potential noise and demonstrate robustness over the training set used.
Unless otherwise specified, we set $n = 4$ and $k = 10$ and use the Adam
optimiser with a learning rate of 0.001 and a L2 penalty (weight decay) of
0.00025 \cite{kingma-ba-2014-adam}. 
We set the minibatch size to 1 as we found that this resulted in a learned
heuristic with the best planning performance and helped the loss function
converge much faster despite the `noisier' training procedure. 
This may be attributed to the small size of our training sets, which is usually
limited to 50-200 samples.

\subsection{Domains and Problems Considered}
The actions in the domains we consider have a unit cost. The problems we train
and evaluate on are randomly generated and unique. We consider the following
domains:

\begin{itemize}
    \setlength\itemsep{0em}
    \item \textbf{8-puzzle} \cite{fern-etal-2011-ipc08}. Our training and test
    set consists of 10 and 50 problems, respectively. Only the initial state
    varies in these problems.
    
    \item \textbf{Sokoban} \cite{fern-etal-2011-ipc08}. Our training set
    consists of 20 problems (10 of grid sizes 5 and 7), and our test set
    contains 50 problems (20 of grid sizes 5 and 7, 10 of size 8).
    The number of boxes was set to 2 and the number of walls randomly selected
    between 3 and 5.
    
    \item
    \textbf{Ferry}\footnote{\url{https://fai.cs.uni-saarland.de/hoffmann/ff-domains.html}}.
    Our training set consists of 9 problems, one for each of the
    parametrisations $\{2, 3, 4 \text{ locations}\}\!\times\!\{1, 2, 3
    \text{ cars}\}$. Our test set contains 36 problems ($\{2, 3, \dots, 10
    \text{ locations}\}\!\times\!\{5, 10, 15, 20 \text{ cars}\}$).
    
    \item \textbf{Blocksworld} \cite{slaney-thiebaux-2001-blocks}. Our training
    set for Blocksworld is formed of 10 problems (5 with 4 and 5 blocks, resp.).
    We have two evaluation sets for  separate experiments: $\mathcal{P}_{bw1}$
    consists of 100 problems (20 with 6,\dots,10 blocks, resp.), while
    $\mathcal{P}_{bw2}$ consists of 50 problems (10 with 4,\dots,8 blocks,
    resp.).
    
    \item \textbf{Gripper} \cite{long-etal-2000-aips}. Our training set for
    Gripper contains 3 Gripper problems (1, 2, 3 balls, resp.). 
    Due to the low number of samples for Gripper (only 20 pairs), we resample
    the training set to 60 samples using stratified sampling with replacement.
    The test set consists of 17 problems with $4,\dots,20$ balls.
    
    \item \textbf{Zenotravel} \cite{long-fox-ippc02}. Our training set consists
    of 10 problems (5 with 2 and 3 cities, resp., with 1-4 planes and 2-5
    people), while our testing set contains 60  problems ($\{2, 3, 4
    \text{cities}\} \times \{2, 3, 4, 5 \text{ planes}\} \times \{3, 4, 5, 6, 7
    \text{ passengers}\}$).
    
\end{itemize}

\subsection{Experimental Results}\label{subsec:results}
Our experiments may be broken down into learning a domain-dependent or a
domain-independent heuristic.
For each of the experiments we describe below, we present the results for the
number of nodes expanded, CPU time, and deviation from the optimal plan length
when using A* (\cref{fig:main-results}).
For \hSpatial{}, the results are presented as the average and its 95\%
confidence interval over the 10 different experiments. 
Additionally, the coverage ratio on the testing problems for each heuristic is
shown in Table \ref{tab:average-coverage}. For \hSpatial{}, we calculate the
average coverage for the 10 repeated experiments.

\begin{table}[t]
\centering
\resizebox{.90\linewidth}{!}{\begin{tabular}{llllll}
                  & Blind & \hMax{} & \hAdd{} & LM-cut & \hSpatial{} \\ \hline
8-puzzle           & 1     & 1     & 1     & 1      & 1         \\ \hline
Sokoban            & 1     & 1     & 1     & 0.96   & 0.91      \\ \hline
Ferry              & 0.42  & 0.36  & 1     & 0.47   & 0.77      \\ \hline
Seen Blocksworld   & 0.78  & 0.68  & 1     & 0.97   & 0.97      \\ \hline
Seen Gripper       & 0.71  & 0.59  & 0.59  & 0.41   & 0.69      \\ \hline
Seen Zenotravel    & 0.62  & 0.55  & 1     & 0.82   & 0.6       \\ \hline
Unseen Blocksworld & 1     & 1     & 1     & 1      & 0.88     
\end{tabular}}
\caption{Coverage Ratio (to 2 d.p.) on the test problems for each heuristic. The
lower average coverage of \hSpatial is attributed our noisy training procedure,
which leads to trained networks with varying performance across experiments.}
\label{tab:average-coverage}
\end{table}

\input{results.tex}

\mypar{Can we learn domain-dependent heuristics?}
In order to evaluate this, we train and test \MyHGNFull{} separately on
8-puzzle, Sokoban and Ferry. 
For 8-puzzle, we limit the training time for each fold to 10 minutes. 
For Sokoban, we use $n = 5$ bins and $k = 5$ folds and limit the training time
within each fold to 20 minutes. 
For Ferry, we use $k = 5$ folds as the training set is quite small (61 samples)
and limit the training time for each fold to 3 minutes.

\cref{fig:8-puzzle,fig:sokoban,fig:ferry} depict the results of these
experiments. 
Firstly, for 8-puzzle, \hSpatial{} expands less nodes than all the baselines
including \hAdd{}, yet \hSpatial{} deviates significantly less from the optimal
plan. 
For Sokoban, \hSpatial{} expands marginally more nodes than \hAdd{} and LM-cut,
but finds near-optimal plans. This is respectable, as Sokoban is known to be
difficult for learning-based approaches as it is PSPACE-complete
\cite{culberson-1997-sokoban}. 
Finally, for Ferry, \hSpatial{} is able to solve problems of much larger size
than the admissible heuristics are able to solve. \hSpatial{} also obtains a
smaller deviation from the optimal than \hAdd{}.
Therefore, \MyHGNFull{} are able to learn domain-dependent heuristics which
potentially outperform our baseline heuristics. 

\mypar{Can we learn domain-independent heuristics?}
To determine whether this is feasible, we train a \MyHGN{} using data from
multiple domains at once: the training set of each domain is binned and
stratified into $k$-folds then, for $i \in \{1,\dots,k\}$, the folds $f_i$ of
all considered domains are merged into a single fold $\hat{f}_i$ and $\hat{F} =
\{\hat{f}_1, \dots, \hat{f}_k\}$ is used as the training set.
Using this procedure, we train a \MyHGNFull{} on the training problems for
Blocksworld, Gripper and Zenotravel; and evaluate the network on the respective
test sets for these domains ($\mathcal{P}_{bw1}$ for Blocksworld). 
We limit the training time for each fold to 15 minutes. Notice that each testing
domain has been \textbf{seen} by the network during training.

\cref{fig:multi-domain-bw,fig:multi-domain-gripper,fig:multi-domain-zeno} depict
our results.
For Blocksworld, \hSpatial{} requires fewer node expansions on average than all
the baselines including \hAdd{}, which compared to \hSpatial{}, deviates
significantly more from the optimal plan length. 
For Gripper, \hSpatial{} requires remarkably less node expansions than the
baselines and is able to find solutions to the larger test problems within the
limited search time (\hMax{} and LM-cut are occluded by \hAdd{} for the number
of nodes expanded).
For Zenotravel, \hSpatial{} requires fewer node expansions than the blind
heuristic, \hMax{} and LM-cut for the more difficult problems. However, at a
certain point we are unable to solve more difficult problems due to the expense
of computing a single heuristic value. Additionally, \hSpatial{} deviates
slightly less from the optimal plan length compared to \hAdd{}.

Thus, \MyHGNFull{} are capable of learning domain-independent heuristics which
generalise to problems from the domains a network has seen during training. 
This is a very powerful result, as current approaches for learning
domain-independent heuristics rely on features derived from existing heuristics,
while we are able to learn heuristics from scratch.

\mypar{Is \hSpatial{} capable of generalising to unseen domains?}
To determine whether this is the case, we train each \MyHGN{} on the training
problems for Zenotravel and Gripper, while we evaluate the network on the test
problems $\mathcal{P}_{bw2}$ for Blocksworld. 
We use the same training data generation procedure described for learning
domain-independent heuristics, and limit the training time for each fold to 10
minutes. 

Notice that Blocksworld is not in the training set, thus it is an
\textbf{unseen} domain for \hSpatial. 
Figure \ref{fig:multi-domain-leave-one-out} depicts the results of this
experiment (one problem for which \hSpatial{} achieved low coverage is left out
as it skews the plots).
We can observe that \hSpatial{} does better than \hMax{} and blind search in
terms of number of node expansions. This is despite the fact that the network
did not see any Blocksworld problems during training.
We note that we ran the experiments with unseen Gripper and unseen Zenotravel
(using the other two domains as the training set). The results for these were
not as promising compared to unseen Blocksworld, but the \MyHGNFull{} still
managed to learn a meaningful heuristic:
for Gripper, the \MyHGNFull{} perform similarly to the admissible heuristics,
including no deviation from the optimal, but do not scale up to large problems;
and for Zenotravel, \MyHGN{} performs better than blind search and \hMax{} but
is outperformed by LM-cut and \hAdd{}.

This shows that it is possible for \hSpatial{} to generalise across to problems
from domains it has not seen during training. 
Unsurprisingly, \hSpatial{} suffers a loss in planning performance in comparison
to networks trained directly on the unseen domain. 

\mypar{Why is \hSpatial{} not competitive in terms of CPU time?}
This may be attributed to our current sub-optimal implementation of
\MyHGNFull{}, and the cost of evaluating the network (i.e., $M$ message passing
steps).
Consequently, there is significant room for improvement in this regard.
Despite this, our results show that \MyHGNFull{} is a feasible and effective
approach for learning domain-dependent and domain-independent heuristics.

\section{Conclusion and Future Work}
We have introduced \MyHGNFull{}, a recurrent encode-process-decode architecture
which uses the Hypergraph Networks framework to learn heuristics which are able
to generalise not only across states, goals, and object sets, but also across
unseen domains. 
In contrast to existing work for learning heuristics, \MyHGNFull{} are able to
learn powerful heuristics from scratch, using only the hypergraph induced by the
delete-relaxation of the STRIPS problem. 
This is achieved by leveraging Hypergraph Networks, which allow us to
approximate optimal heuristic values by performing message passing on features
in a rich latent space.
Our experimental results show that \MyHGNFull{} are able to learn
domain-dependent and domain-independent heuristics which are competitive with 
\hMax{}, \hAdd{} and LM-cut, which are computed over the same hypergraph, in
terms of the number of node expansions required by A*. 
This suggests that learning heuristics over hypergraphs is a promising approach,
and hence should be investigated in further detail.

Potential avenues for future work include using a richer set of input features
such as the
\textit{disjunctive action landmarks}
or the \textit{fact landmarks}.
This may help the network learn a heuristic which is closer to the optimal and
reduce the number of message passing steps required to obtain an informative
heuristic estimate.
Moreover, the time required to compute a single heuristic value ($\approx$0.01
to 0.02 seconds) could be reduced significantly by optimising our implementation
(e.g., using multiple CPU cores or GPUs, optimising matrix operations and
broadcasting), adapting the number of message passing steps in real time, or
even pruning the vertices and hyperedges in the hypergraph of the relaxed
problem.

Finally, we may investigate how to adapt \MyHGNFull{} for Stochastic Shortest
Path problems (SSPs)~\cite{bertsekas-1991-ssp}.
Existing heuristics for SSPs either rely on linear
programming~\cite{trevizan-etal-2017-occupation}, which can be expensive, or
rely on \textit{determinisation}, which oversimplifies the probabilistic
actions.
It may be possible to use Hypergraph Networks to learn an informative heuristic
that preserves the probabilistic structure of actions by deriving suitable
hypergraphs from \textit{factored} SSPs. 

\bibliographystyle{aaai}
\bibliography{references}

\end{document}

%% file: introduction.tex
\section{Introduction}
Despite the prevalence of deep learning for perception tasks in computer vision
and natural language processing, its application to problem solving tasks, such
as planning, is still in its infancy. The majority of deep learning approaches
to planning use conventional architectures designed for perception tasks, rely
on hand-engineering features or encoding planning problems as images, and do not
learn knowledge that generalises beyond planning with a different initial state
or goal
\cite{buffet:aberdeen:09,arfaee-etal-2010-bootstrap,groshev-etal-2018l-grp}.
One exception is Action Schema Networks (ASNets)
\cite{toyer:etal:18,toyer-etal-2019-asnets}, a neural network architecture which
exploits the relational structure of a given planning domain described in
(P)PDDL, to learn generalised policies applicable to problems of any size within
the domain.

The motivation of our work is to go even further than architectures such as
ASNets, and learn to plan -- or at least to guide the search for a plan --
independently of the domain considered.
In particular, we consider the problem of learning domain-independent heuristics
that generalise not only across states, goals, and object sets, but also across
domains.

We focus on the well-known class of delete-relaxation heuristics for
propositional STRIPS planning
\cite{bonet-geffner-2001-search,helmert-domshlak-2009-lmcut}, of which
$h^{max}$, $h^{add}$, and LM-cut are popular examples.
These heuristics can be seen as the least-cost path in the hypergraph
representing the delete-relaxed problem for a suitable aggregation function.
The vertices of this hypergraph represent the problem's propositions and the
hyperedges represent actions connecting their preconditions to their positive
effects.
We can therefore frame the problem of learning domain-independent heuristics as
that of learning a mapping from the hypergraph representation of the
delete-relaxed problem (and optionally other features) to a cost estimate. 
To develop and evaluate this hypergraph learning framework, we make three
contributions:

\begin{enumerate} \setlength\itemsep{0em}
\item \textbf{Hypergraph Networks (HGNs)}, our novel framework which generalises
Graph Networks \cite{battaglia-etal-2018-relational} to hypergraphs. The HGN
framework may be used to design new hypergraph deep learning models, and
inherently supports combinatorial generalisation to hypergraphs with different
numbers of vertices and hyperedges.
\item \textbf{\MyHGNFull{}}, an instance of a HGN which is designed to learn
heuristics by
approximating shortest paths over the hypergraph induced by the delete
relaxation of a STRIPS problem.  \MyHGNFull{} use a powerful recurrent
encode-process-decode architecture which allows them to incrementally propagate
messages within the hypergraph in latent space.
\item \textbf{A detailed empirical evaluation}, which rigorously defines the
Hypergraph Network configurations and training procedure we use in our
experiments. We train and evaluate our \MyHGNFull{} on a variety of domains and
show that they are able learn domain-dependent and domain-independent heuristics
which potentially outperform $h^{max}$, $h^{add}$, and LM-cut.
\end{enumerate} 
As far as we are aware, this is the first work to learn domain-independent
heuristics completely from scratch.

\section{Related Work}
There is a large body of literature on learning for planning. Jimenez et al.
\shortcite{jimenez:etal:12} and Toyer et al. \shortcite{toyer-etal-2019-asnets}
provide excellent surveys on these existing approaches. 
Due to space limitations, we focus on deep learning (DL) approaches to planning
which differ in what they learn, the features and architectures they use, and
the generality they confer.

\mypar{What is learned?} 
Existing DL approaches may be split into four categories: learning {\em domain
descriptions} \cite{say:etal:17,asai:fukunaga:18}, 
{\em policies} \cite{buffet:aberdeen:09,toyer:etal:18,groshev-etal-2018l-grp,issakkimuthu-fern-tadepalli-2018-drp,garg-bajpai-mausam-2019-rddl},
{\em heuristics}
\cite{samadi:etal:08,arfaee-etal-2010-bootstrap,thayer:etal:11,gomoluch-etal-2017-heuristics},
and {\em planner selection} \cite{sievers-etal-2019-dl}. 
Our work is concerned with learning heuristics. 
One of the key differences of our approach with the existing state-of-the-art
for learning heuristics is that we learn heuristics from scratch instead of
improving or combining existing heuristics.
That being said, \MyHGNFull{}\ are also suitable to learn heuristic improvements
or combinations, and with some adaptations, to learn actions rankings; however,
we have not experimented with these settings.

\mypar{Features and Architectures.}
Most existing DL approaches to planning use standard architectures, and rely on
hand-engineered features or encodings of planning problems as images.
For instance, Sievers et al. \shortcite{sievers-etal-2019-dl} train
Convolutional Neural Networks (CNNs) over graphical representations of planning
problems converted into images, to determine which planner should be invoked for
a planning task. 
For learning generalised policies and heuristics, Groshev et al.
\shortcite{groshev-etal-2018l-grp} train CNNs and Graph Convolutional Networks
with images obtained via a domain-specific hand-coded problem conversion.
In contrast, our approach does not require hand-coded features and instead
learns latent features directly from a rich hypergraph representation of the
planning problem. 

Another approach is ASNets \cite{toyer:etal:18}, a neural network architecture
dedicated to planning, composed of alternating action and proposition layers
which are sparsely connected according to the relational structure of the action
schemas in a (P)PDDL domain. 
A disadvantage of ASNets is its fixed receptive field which limits its
capability to support long chains of reasoning. Our \MyHGNFull{}\ architecture
does not have such an intrinsic receptive field limitation.

\mypar{Generalisation.}
Existing approaches and architectures for learning policies and heuristics have
limited generalisation capabilities. 
Many generalise to problems with different initial states and goals, but not to
problems with different sets or numbers of objects.  
Exceptions include ASNets, whose weight sharing scheme allows the generated
policies to generalise to problems of any size from a given (P)PDDL domain, and
\textsc{TraPSNet} \cite{garg-bajpai-mausam-2019-rddl}, whose graph attention
network can be transferred between different numbers of objects in an RDDL
domain. 
As our experiments show, not only does \MyHGNFull{}\ support generalisation
across problem sizes, but it also supports learning domain-independent
heuristics that generalise across domains, including to domains that were not
seen during training.

%% file: hypergraph-networks.tex
\section{Hypergraph Networks}
Hypergraph Networks (HGNs) is our generalisation of the Graph Networks
\cite{battaglia-etal-2018-relational} framework to hypergraphs. 
HGNs may be used to represent and extend existing DL models including CNNs,
graph neural networks, and state-of-the-art hypergraph neural networks. 
We will not explore HGNs in great detail, as it is not the focus of this paper. 

\begin{figure}[t]
    \centering
    \includegraphics[width=.65\linewidth]{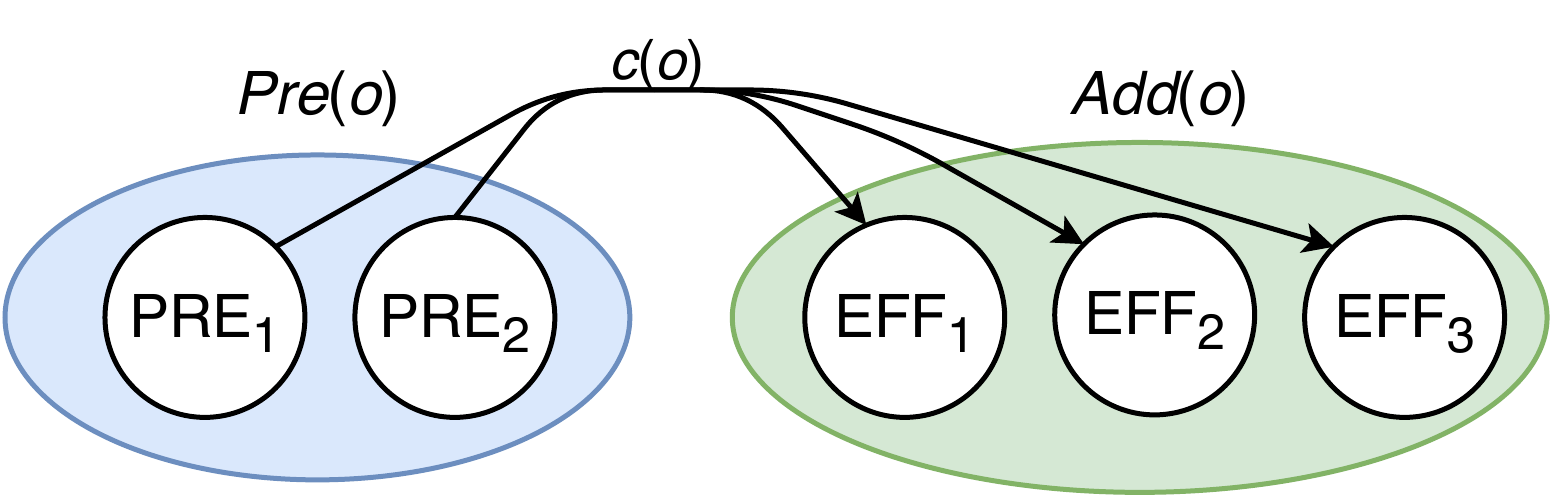}
    \caption{Our formulation of a hyperedge for an action $o \in O$ with 2
    preconditions and 3 positive effects. The preconditions are the `senders',
    while the positive effects are the `receivers'.}
    \label{fig:strips-hyperedge}
\end{figure}

\mypar{Hypergraph Definition.}
A hypergraph is a generalisation of a graph in which a hyperedge may connect any
number of vertices together.
A directed hypergraph in the HGN framework is defined as a triple $G =
(\mathbf{u}, V, E)$ where:
$\mathbf{u}$ represents the hypergraph-level (global) features;
$V = \{\mathbf{v}_i\ \colon i \in \{1, \dots, N^v\}\}$ is the set of $N^v$
vertices where $\mathbf{v}_i$ represents the $i$-th vertex's features;
and $E = \{(\mathbf{e}_k, R_k, S_k) \colon k \in \{1, \dots, N^e\}\}$ is the set
of $N^e$ hyperedges, where $\mathbf{e}_k$ represents the $k$-th hyperedge's
features, $R_k$ is the vertex set which contains the indices of the vertices
which are in the head of the $k$-th hyperedge (i.e., receivers), and $S_k$ is
the vertex set which contains the indices of the vertices which are in the tail
of the $k$-th hyperedge (i.e., senders). 
This is in contrast to Graph Networks, where $R_k$ and $S_k$ are singletons,
i.e., $\vert R_k \vert = \vert S_k \vert = 1$. 
An example of a hyperedge for a delete-relaxed STRIPS action is depicted in
\cref{fig:strips-hyperedge}.

\mypar{Hypergraph Network Block.}
A Hypergraph Network (HGN) block is a hypergraph-to-hypergraph function which
forms the core building block of a HGN. 
The internal structure of a HGN block is identical to a Graph Network block
\cite{battaglia-etal-2018-relational}, except now the hyperedge update function
$\phi^e$ supports multiple receivers and senders. 
A \textit{full} HGN block is composed of three update functions, $\phi^e$,
$\phi^v$ and $\phi^u$, and three aggregation functions, $\rho^{e \rightarrow
v}$, $\rho^{e \rightarrow u}$ and $\rho^{v \rightarrow u}$:
\begin{equation*}
  \begin{split}
    &\mathbf{e}'_k = \phi^e(\mathbf{e}_k, \mathbf{R}_k, \mathbf{S}_k, \mathbf{u}) \\
    &\mathbf{v}'_i = \phi^v(\overline{\mathbf{e}}'_i, \mathbf{v}_i, \mathbf{u}) \\
    &\mathbf{u}' = \phi^u(\overline{\mathbf{e}}', \overline{\mathbf{v}}', \mathbf{u})
  \end{split}
\qquad\qquad
  \begin{split}
    &\overline{\mathbf{e}}'_i = \rho^{e \rightarrow v} (E'_i) \\
    &\overline{\mathbf{e}}' = \rho^{e \rightarrow u} (E') \\
    &\overline{\mathbf{v}}' = \rho^{v \rightarrow u} (V')
  \end{split}
\end{equation*}
where $\mathbf{R}_k = \{\mathbf{v}_j \colon j \in R_k\}$ and $\mathbf{S}_k =
\{\mathbf{v}_j \colon j \in S_k\}$ are the sets which represent the vertex
features of the receivers and senders of the $k$-th hyperedge, respectively.
Additionally, for the $i$-th vertex, we define $E'_i = \{(\mathbf{e}'_k, R_k,
S_k) \colon k \in \{1, \dots, N^e\} \text{ s.t. } i \in R_k\}$, $V' =
\{\mathbf{v}'_i \colon i \in \{1, \dots, N^e\} \}$, and $E' = \bigcup_i E'_i =
\{(\mathbf{e}'_k, R_k, S_k) \colon k \in \{1, \dots, N^e\}\}$. 
Essentially, $E'_i$ represents the updated hyperedges where the $i$-th vertex is
a receiver vertex, $E'$ represents all the updated hyperedges, and $V'$
represents all the updated vertices. 

Since the input to the aggregation functions are essentially sets, each $\rho$
must be permutation invariant to ensure that all permutations of the input give
the same aggregated result. 
Hence $\rho$ could, for example, be a function that takes an element-wise
summation of the input, maximum, minimum, mean, etc
\cite{battaglia-etal-2018-relational}. 

\mypar{Computation Steps.} In a single forward pass of a HGN block, the
hyperedge update function $\phi^e$ is firstly applied to all hyperedges to
compute per-hyperedge updates. 
Each updated hyperedge feature $\mathbf{e}'_k$ is computed using the current
hyperedge's feature $\mathbf{e}_k$, the features of the receiver and sender
vertices $\mathbf{R}_k$ and $\mathbf{S}_k$, respectively, and the global
features $\mathbf{u}$.
Next, the vertex update function $\phi^v$ is applied to all vertices to compute
per-vertex updates. 
Each updated vertex feature $\mathbf{v}'_i$ is computed using the aggregated
information $\overline{\mathbf{e}}_i'$ from all the hyperedges the vertex
`receives' a signal from (i.e., it appears in the head of the hyperedge), the
current vertex's feature $\mathbf{v}_i$, and the global features $\mathbf{u}$.
Finally, the global update function $\phi^u$ is applied to compute the new
global features~\cite{battaglia-etal-2018-relational} using the aggregated
information from all the hyperedges and vertices in the hypergraph
$\overline{\mathbf{e}}'$ and $\overline{\mathbf{v}}'$, respectively, along with
the current global features $\mathbf{u}$. 

\begin{figure}[t]
    \centering
    \includegraphics[width=.75\linewidth]{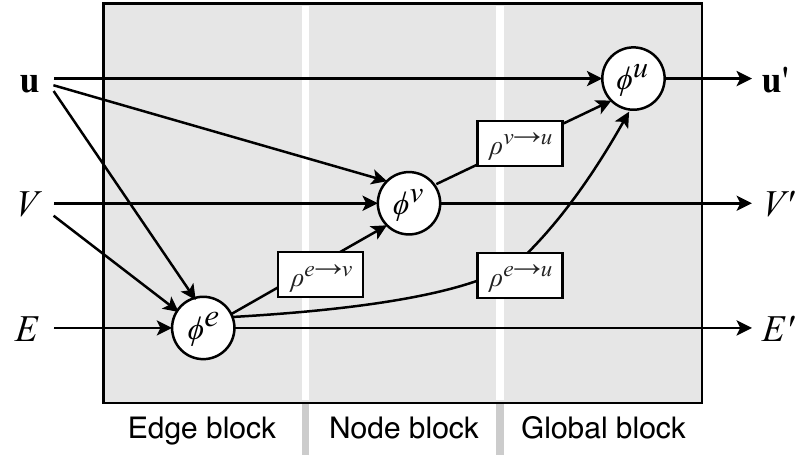}
    \caption{The full HGN block configuration which predicts global, vertex and
    hyperedge outputs based on the incoming global, vertex and hyperedge
    attributes (Figure 4a from \citealp{battaglia-etal-2018-relational}). The
    incoming arrows to an update function $\phi$ represent the inputs it
    receives.}
    \label{fig:hgn-block}
\end{figure}

\mypar{Configuring HGN Blocks.}
Each update function $\phi$ in a HGN block must be implemented by some function
$f$, where the signature of $f$ determines what input it gets
\cite{battaglia-etal-2018-relational}. 
For example, the function that implements $\phi^e$ in a full HGN block
(\cref{fig:hgn-block}) is a function $f \colon (\mathbf{e}_k, \mathbf{R}_k,
\mathbf{S}_k, \mathbf{u}) \mapsto \mathbf{e}'_k$ which accepts the global,
vertex, and hyperedge attributes. 
Each function $f$ may be implemented in any manner, as long as it accepts the
input parameters and conforms to the required output.
Since HGN blocks are hypergraph-to-hypergraph functions, we may compose blocks
sequentially and repeatedly apply them.

%% file: strips-hgn.tex
\section{\MyHGNFull{}{}}
\label{sec:strips-hgn}
\MyHGNFull{} is our instantiation of a HGN which uses a recurrent
\textit{encode-process-decode} architecture
\cite{battaglia-etal-2018-relational} for learning heuristics. 
\MyHGNFull{} are designed to be highly adaptable to different input features for
each proposition and action, as well as being agnostic to the implementation of
each update function in each HGN block.

\mypar{Hypergraph Representation.} The input to a \MyHGN{} is a hypergraph
$G_{\text{inp}} = (\mathbf{u}_{\text{inp}}, V_{\text{inp}}, E_{\text{inp}})$
which contains the input proposition and action features for the state $s$,
along with the hypergraph structure of the relaxed STRIPS problem $P^+ = \langle
F, O', I, G, c \rangle$, where:
\begin{itemize} \setlength{\itemsep}{2pt} \setlength{\parskip}{0pt}
    \item $\mathbf{u}_{\text{inp}} = \emptyset$, as global features are not
    required as input to a \MyHGN{}. 
    Nevertheless, it is easy to adapt \MyHGNFull{} to support global features,
    e.g., we could supplement a \MyHGN{} with a heuristic value $h(s)$ computed
    by another heuristic $h$ such that the network learns an ``improvement'' on
    $h$ similar to~\cite{gomoluch-etal-2017-heuristics}.
    \item  $V_{\text{inp}} = \{\mathbf{v}_i\ \colon i \in \{1, \dots, \lvert F
    \rvert \}\}$ contains the input features for the $\lvert F \rvert$
    propositions in the problem. 
    Features for a proposition could include whether it is true for the current
    state or goal state, and whether the proposition is a \textit{fact landmark}
    for the state $s$~\cite{richter-westphal-2010-lama}.
    \item $E_{\text{inp}} = \{(\mathbf{e}_k, R_k, S_k) \colon k \in \{1, \dots,
    \lvert O' \rvert \}\}$ for the $\lvert O' \rvert$ actions in the relaxed
    problem $P^+$. 
    For an action $o \in O'$ represented by the $k$-th hyperedge, $\mathbf{e}_k$
    represents the input features for $o$ (e.g., the cost of the action $c(o)$,
    and whether the action is in the \textit{disjunctive action landmarks} from
    state $s$) and $R_k = Add(o)$ (resp. $S_k = Pre(o)$) is the vertex set
    containing the indices of the vertices in the additive effects (resp.
    preconditions) of $o$.
    
\end{itemize}

The output of a \MyHGN{} is a hypergraph $G_{\text{out}} =
(\mathbf{u}_{\text{out}}, V_{\text{out}}, E_{\text{out}})$ where
$\mathbf{u}_{\text{out}} \in \mathbb{R}^{1 \times 1}$ is a 1-dimensional vector
representing the heuristic value for $s$, thus we enforce both $V_{\text{out}}$
and $E_{\text{out}}$ to be the empty set.

\subsection{Architecture}
\begin{figure}[t]
    \centering
    \includegraphics[width=.85\linewidth]{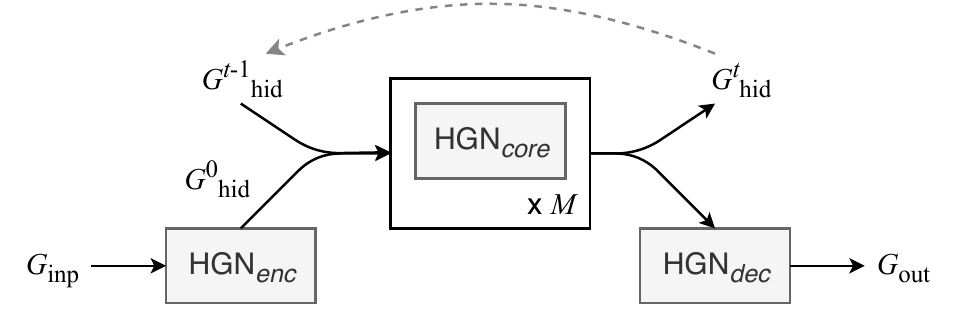}
    \caption{The recurrent encode-process-decode architecture of a \MyHGN{} 
    (modified from Figure 6c in \citealp{battaglia-etal-2018-relational}). The
    merging line for $G^0_{\text{hid}}$ and $G^{t-1}_{\text{hid}}$ indicates
    concatenation, while the splitting lines that are output by the
    $\text{HGN}_{core}$ block indicate copying (i.e., the same output is passed
    to different locations). The grey dotted line indicates that the output
    $G^t_{\text{hid}}$ is used as input to the $\text{HGN}_{core}$ block in the
    next time step $t + 1$.}
    \label{fig:hgn-enc-proc-dec}
\end{figure}

A \MyHGN{} is composed of three main HGN blocks: the encoding, processing
(core), and decoding block. 
Our architecture follows a recurrent \textit{encode-process-decode} design
\cite{hamrick-etal-2018-relational,battaglia-etal-2018-relational}, as depicted
in \cref{fig:hgn-enc-proc-dec}.
The input hypergraph $G_{\text{inp}}$ is firstly encoded to a latent
representation $G^0_{\text{hid}}$ by the encoding block $\text{HGN}_{enc}$ at
time step $t = 0$. 
This allows the network to operate on a richer representation of the input
features in latent space.

Next, the initial latent representation of the hypergraph $G^0_{\text{hid}}$ is
concatenated with the previous output of the processing block
$\text{HGN}_{core}$. 
Initially, when $\text{HGN}_{core}$ has not been called (i.e., at time step $t =
1$ just after $G_{\text{inp}}$ has been computed), $G^0_{\text{hid}}$ is
concatenated with itself. 
Note that the hypergraph structure for $G^0_{\text{hid}}$ and
$G^{t-1}_{\text{hid}}$ is identical because the HGN blocks do not update the
senders or receivers for a hyperedge.
Implementation-wise, concatenating a hypergraph with another involves
concatenating the features for each corresponding vertex $\mathbf{v}_i$
together, and the features for each corresponding hyperedge $\mathbf{e}_k$
together (the global features are not concatenated as they are not required as
input to a \MyHGN{}). 
This results in a broadened feature vector for each vertex and hyperedge.

The processing block $\text{HGN}_{core}$, which outputs a hypergraph
$G^t_{\text{hid}}$ for each time step $t \in \{1, \dots, M\}$, is applied $M$
times with the initial encoded hypergraph $G^0_{\text{hid}}$ concatenated with
the previous output of $\text{HGN}_{core}$ as the input (see
\cref{fig:hgn-enc-proc-dec}). 
Evidently, this results in $M\!-\!1$ intermediate hypergraph outputs, one for
each for time step $t \in \{1, \dots, M\!-\!1\}$, and one final hypergraph for
the time step $t = M$.  
The decoding block takes the hypergraph output by the $\text{HGN}_{core}$ block
and decodes it to the hypergraph $G_{\text{out}}$ which contains the heuristic
value for state $s$ in the global feature $\mathbf{u}_{\text{out}}$. 
Observe that we can decode each latent hypergraph which is output by
$\text{HGN}_{core}$ to obtain a heuristic value for each time step $t \in \{1,
\dots, M\}$. 
We use this fact to train a \MyHGN{} by optimising the loss on the output of
each time step.

\begin{figure*}[t!]
\centering
\vspace*{-4mm}
\begin{subfigure}[t]{0.28\textwidth}
  \centering
  \includegraphics[width=\linewidth]{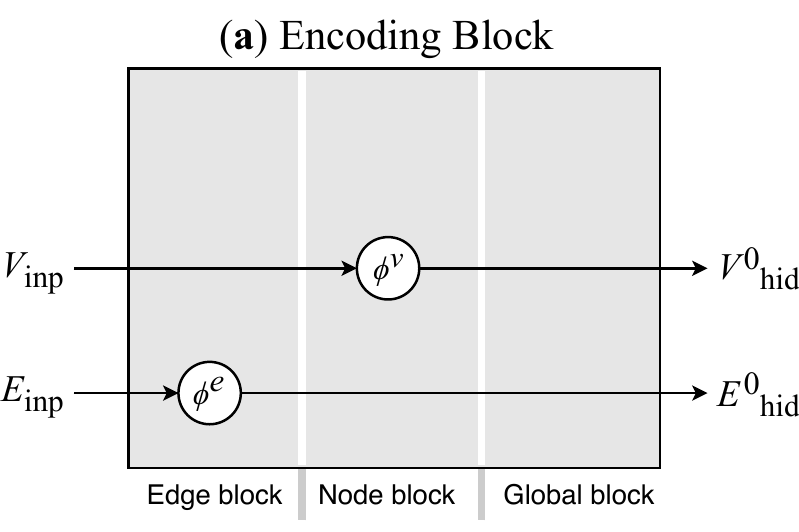}
  \phantomcaption  \label{fig:hgn-enc-block}
\end{subfigure}%
~ ~ ~ ~ ~ 
\begin{subfigure}[t]{0.39\textwidth}
  \centering
  \includegraphics[width=\linewidth]{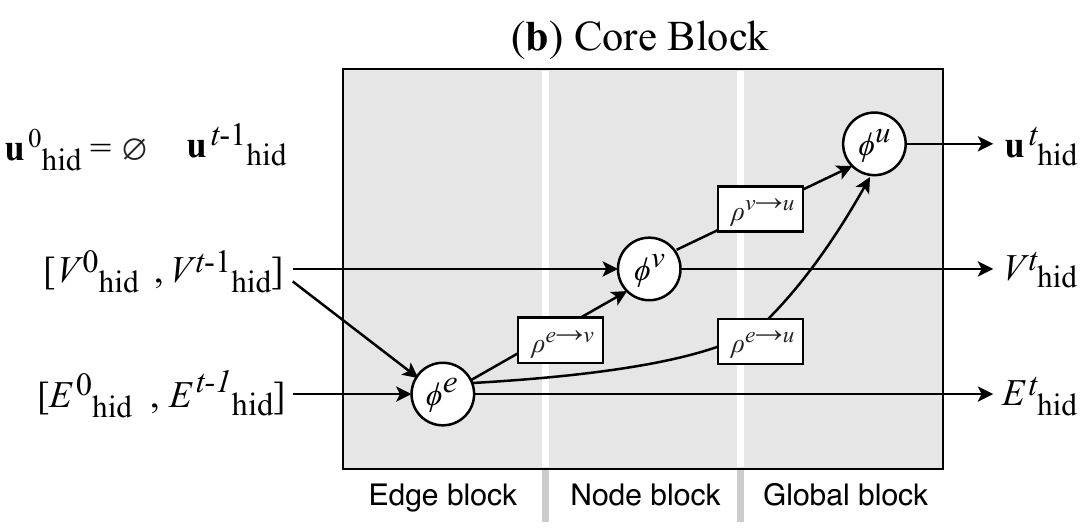}
  \phantomcaption \label{fig:hgn-core-block}
\end{subfigure}%
~ ~ ~ ~ ~
\begin{subfigure}[t]{0.28\textwidth}
  \centering
  \includegraphics[width=\linewidth]{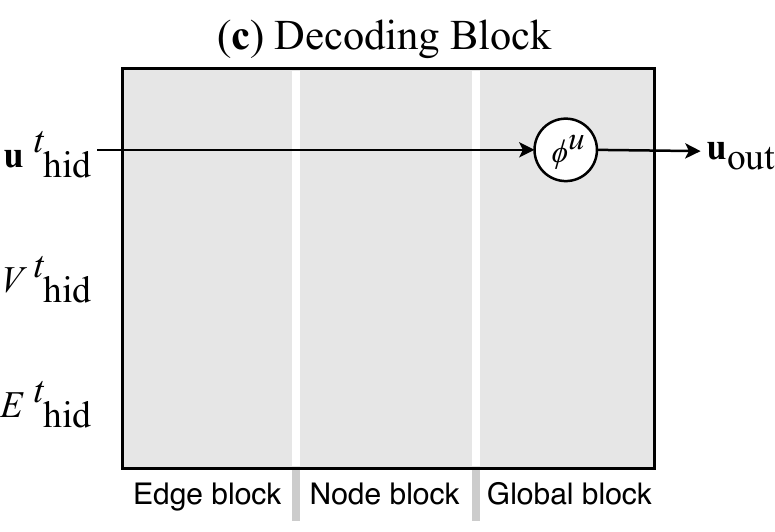}
  \phantomcaption \label{fig:hgn-dec-block}
\end{subfigure}%
\vspace*{-4mm}
\caption{%
The encoding (a), core (b) and decoding (c) blocks of a \MyHGN{}.
The encoding block independently encodes the vertex and hyperedge features into
latent space.
The core block computes per-hyperedge and per-vertex updates using the
concatenated input hypergraph, and additionally computes a latent heuristic
feature $\mathbf{u}^t_{\text{hid}}$.
The decoding block decodes the latent heuristic features
$\mathbf{u}^t_{\text{hid}}$ into a single heuristic value.}
\end{figure*}

\mypar{Core Block Details.}
We can interpret a \MyHGN{} as a message passing model which performs $M$ steps
of message passing \cite{gilmer-etal-2017-mpnn}, as the shared processing block
$\text{HGN}_{core}$ is repeated $M$ times using a recurrent architecture. 
A single step of message passing is equivalent to sending a `signal' from a
vertex to its immediate neighbouring vertices. Although this means that a vertex
only receives a `signal' from other vertices at most $M$ hops away,
%
%
we theorise that this is sufficient to learn a powerful function which
aggregates proposition and action features in the latent space.

In contrast to architectures such as ASNets and CNNs, which have a fixed
receptive field that is determined by the number of hidden layers, the receptive
field of a \MyHGN{} is effectively determined by the number of message passing
steps. 
Evidently, we can increase or decrease the receptive field of a \MyHGN{} by
scaling the number of message passing steps, hence providing a significant
advantage over networks with fixed receptive fields.

\mypar{Within-Block Design.}
The encoder block (\cref{fig:hgn-enc-block}) $\text{HGN}_{enc}$ encodes the
vertex and hyperedge input features independently of each other using its
$\phi^v$ and $\phi^e$, respectively. 

The core processing block of a \MyHGN{} (\cref{fig:hgn-core-block}) takes the
concatenated vertex and hyperedge features from the latent hypergraphs
$G^0_{\text{hid}}$ and $G^{t-1}_{\text{hid}}$ as input. 
$\phi^e$ computes per-hyperedge updates based on these hyperedge and vertex
features. 
$\phi^v$ computes per-vertex updates based on the vertex features and the
aggregated features of the hyperedges where the vertex is a receiver, which is
computed using $\rho^{e \rightarrow v}$. 
Finally, $\phi^u$ uses the aggregated vertex and aggregated hyperedge features
calculated with $\rho^{e \rightarrow v}$ and $\rho^{e \rightarrow u}$,
respectively, to compute a latent representation for the heuristic value.

The decoder block (\cref{fig:hgn-dec-block}) takes the latent representation of
the global features $\mathbf{u}^t_{\text{hid}}$ of the hypergraph returned by
the core HGN block and uses its $\phi^u$ to decode it into a one-dimensional
heuristic value. 
The vertex and hyperedge features are not used as $\mathbf{u}^t_{\text{hid}}$
already represents an aggregation of these features as computed by
$\text{HGN}_{core}$.

The choice of learning model for the update functions $\phi^e$, $\phi^v$ and
$\phi^u$ within each block is not strict, as long as the model conforms to the
input and output requirements. 
The choice of aggregation functions $\rho^{e \rightarrow v}$, $\rho^{e
\rightarrow u}$, and $\rho^{v \rightarrow u}$ should be permutation invariant to
the ordering of the inputs, otherwise different heuristic values could be
obtained for different permutations of the same STRIPS problem.
We detail our choice of update and aggregation functions in Section
\ref{subsec:hgn-conf-exp} which describes our experimental setup.

\subsection{Training Algorithm}\label{subsec:training}
We consider learning a heuristic function $h$ as a regression problem, where $h$
ideally provides near-optimal estimates of the cost to go.
We train our \MyHGNFull{} with the values generated by the optimal heuristic
$h^*$. Given a set of training problems $\mathcal{P} = \{P_1, \dots, P_n\}$, we
run an optimal planner for each $P_i \in \mathcal{P}$ to obtain optimal
state-value pairs $(s, h^*(s))$. 
We then generate the delete-relaxed hypergraph $G$ for $P_i$ and the state $s$
to get a training sample $(G, h^*(s))$.
We denote by $\mathcal{T}$ the set containing all training samples.

\mypar{Weight Optimisation.}
We use supervised learning and assume that each update function in the encoder,
core, and decoder blocks of a \MyHGN{} has some weights that need to be learned.
For simplicity, we aggregate these weights into a single variable $\theta$.
Let $h^{\theta}$ be the heuristic learned by a \MyHGN{} which is parameterised
by the weights $\theta$.

Recall that we can decode the latent hypergraph that is output by the core HGN
block at each time step $t \in \{1, \dots, M\}$ into a heuristic value
$h^{\theta}_t$. 
Our loss function averages the losses of these intermediate outputs at each time
step to encourage a \MyHGN{} to find a good heuristic value in the smallest
number of message passing steps possible \cite{battaglia-etal-2018-relational}.
We use the mean squared error (MSE) loss function:
\[
    \mathcal{L}_{\theta}(\mathcal{B}) = \frac{1}{\lvert \mathcal{B}\rvert}\!
    \sum_{(G, \ h^*(s)) \in \mathcal{B}}\!\frac{1}{M} \!
    \sum_{t \in \{1, \dots, M\}}\!\left(h^{\theta}_t(G) - h^*(s) \right)^2
\]
\noindent where $\mathcal{B} \subseteq \mathcal{T}$ is a minibatch within the
entire training set $\mathcal{T}$, $M$ is the number of message passing steps,
and $G$ is the input hypergraph for state $s$ in a problem. 

We use minibatch gradient descent \cite{li-etal-2014-minibatch} to update the
weights $\theta$ in the direction which minimises $L_\theta$ by using the
gradient $d\mathcal{L}_{\theta}(\mathcal{B}) / d\theta$. 
In a single epoch, we apply this update to every minibatch $\mathcal{B}$. 
We repeatedly apply more epochs until we reach a maximum number of epochs or
exceed a fixed training time.
During evaluation time, we use the heuristic value $h^\theta_M$ output at the
last message step $t = M$. 

\subsection{Limitations of \MyHGNFull{}}
Firstly, it is expensive to compute a single heuristic value using a \MyHGN{},
given the computational cost of the matrix operations required for a single step
of message passing; these costs scale with the number of vertices and hyperedges
in the hypergraph. 
However, this cost may pay off if the learned heuristic provides very
informative estimates near the optimal $h^*$, as it may reduce the total CPU
time required to find a near-optimal solution.

The number of message passing steps $M$ for the core HGN block is a
hyperparameter which, in theory, should be adaptively selected based on how
`far' away the current state is from the goal.
However, determining a good value for $M$ is not trivial, and should ideally be
automatically determined by a \MyHGN{} by using its intermediate outputs.
In practice, we found that setting $M\!=\!10$ was sufficient to achieve
promising results.

Finally, we are unable to provide any formal guarantees that the heuristics
learned by \MyHGNFull{} are admissible. Although we train \MyHGNFull{} on the
optimal heuristic values, it is unfeasible to analyse a network to understand
what it is exactly computing.

%% file: results.tex
\clearpage
\newcommand{\plotlength}{.95\textwidth}
\newcommand{\trimlength}{7mm}
\newcommand{\plottitle}[1]{{\footnotesize #1}}
\newcolumntype{C}[1]{>{\centering\let\newline\\\arraybackslash\hspace{0pt}}m{#1}}
\begin{figure*}[t]
{\footnotesize\begin{tabular}{%
p{0.025\textwidth}%
C{0.30\textwidth}%
C{0.285\textwidth}%
C{0.28\textwidth}%
p{0.01\textwidth}%
}
& \textbf{Number of Nodes Expanded by A*} &
\textbf{Total CPU Time for A*} &
\textbf{Deviation from Optimal Plan Length} &
\end{tabular}}\\[2mm]
\centering
\plottitle{(a) 8-Puzzle}
\begin{subfigure}[t]{\textwidth}
  \centering
  \includegraphics[trim={0 0 0 \trimlength},clip,width=\plotlength]{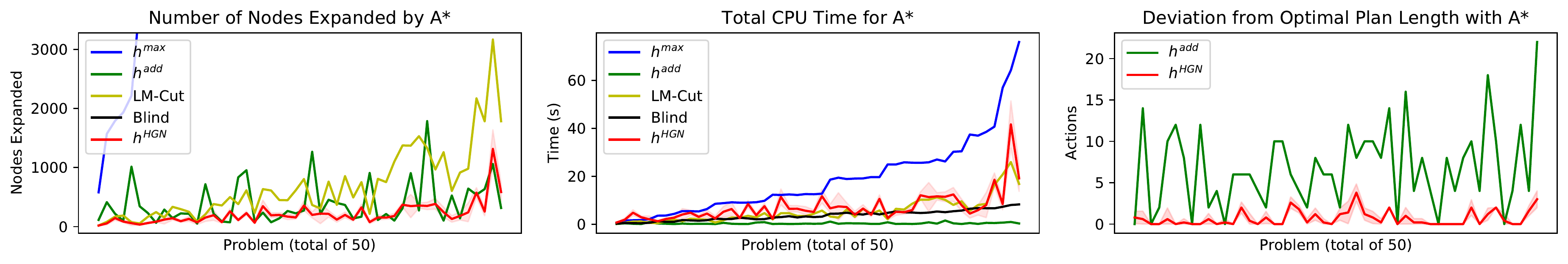}
  \phantomcaption \label{fig:8-puzzle}
\end{subfigure}\\
\plottitle{(b) Sokoban}
\begin{subfigure}[t]{\textwidth}
  \centering
  \includegraphics[trim={0 0 0 \trimlength},clip,width=\plotlength]{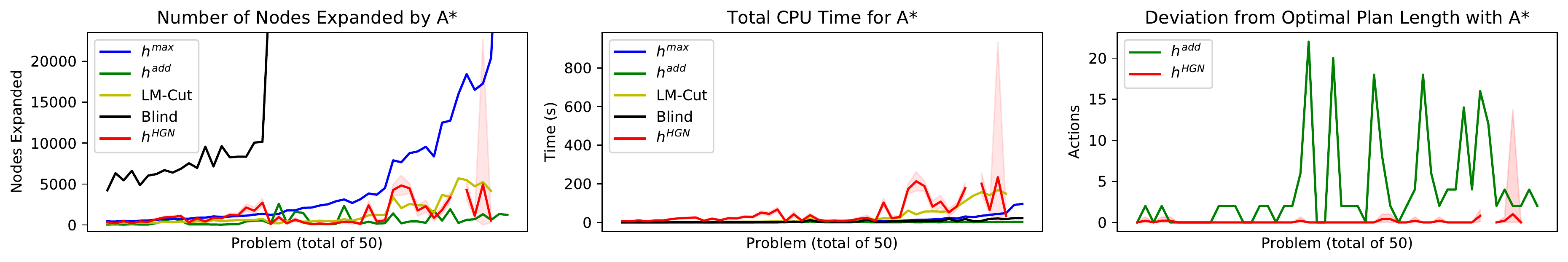}
  \phantomcaption \label{fig:sokoban}
\end{subfigure}\\
\plottitle{(c) Ferry}
\begin{subfigure}[t]{\textwidth}
  \centering
  \includegraphics[trim={0 0 0 \trimlength},clip,width=\plotlength]{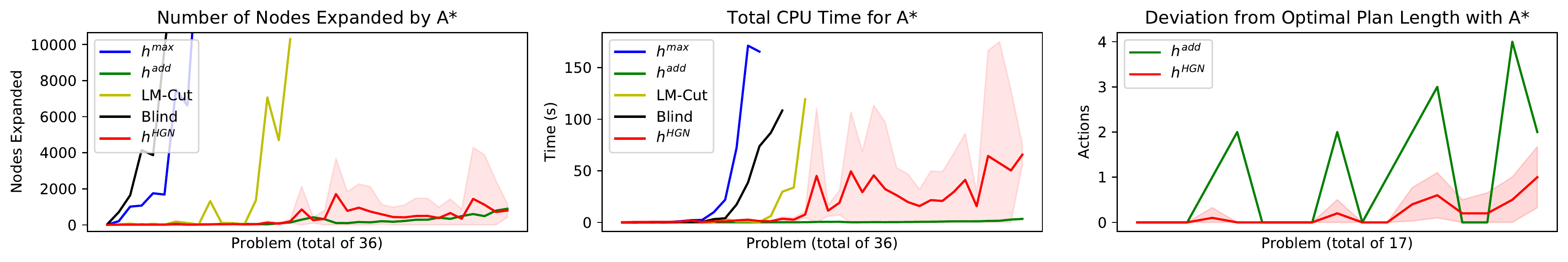}
  \phantomcaption \label{fig:ferry}
\end{subfigure}\\
\plottitle{(d) Domain-independent Blocksworld}
\begin{subfigure}[t]{\textwidth}
  \centering
  \includegraphics[trim={0 0 0 \trimlength},clip,width=\plotlength]{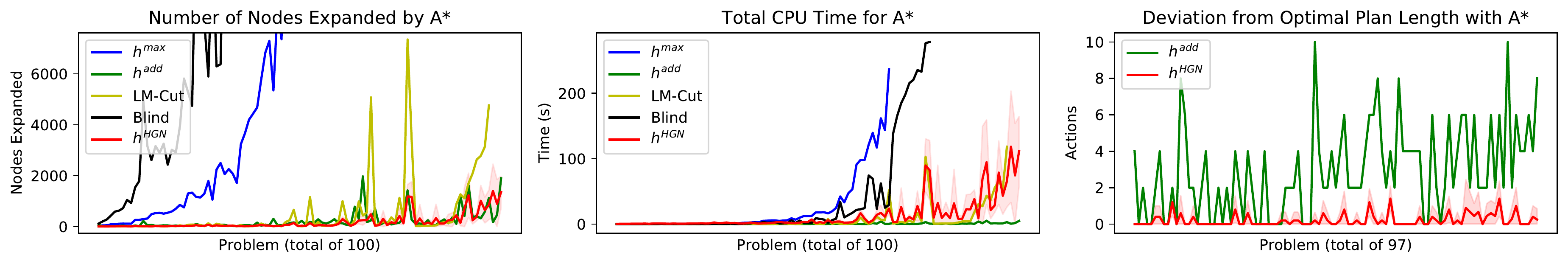}
  \phantomcaption \label{fig:multi-domain-bw}
\end{subfigure}\\
\plottitle{(e) Domain-independent Gripper}
\begin{subfigure}[t]{\textwidth}
  \centering
  \includegraphics[trim={0 0 0 \trimlength},clip,width=\plotlength]{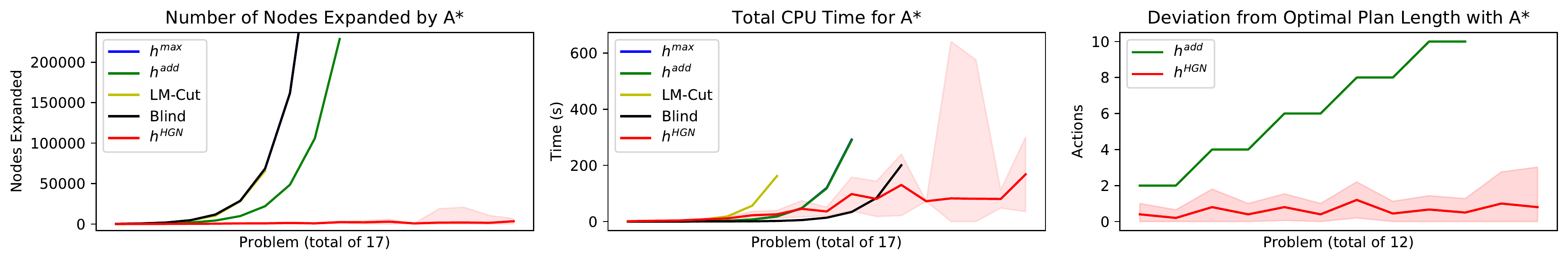}
  \phantomcaption \label{fig:multi-domain-gripper}
\end{subfigure}\\
\plottitle{(f) Domain-independent Zenotravel}
\begin{subfigure}[t]{\textwidth}
  \centering
  \includegraphics[trim={0 0 0 \trimlength},clip,width=\plotlength]{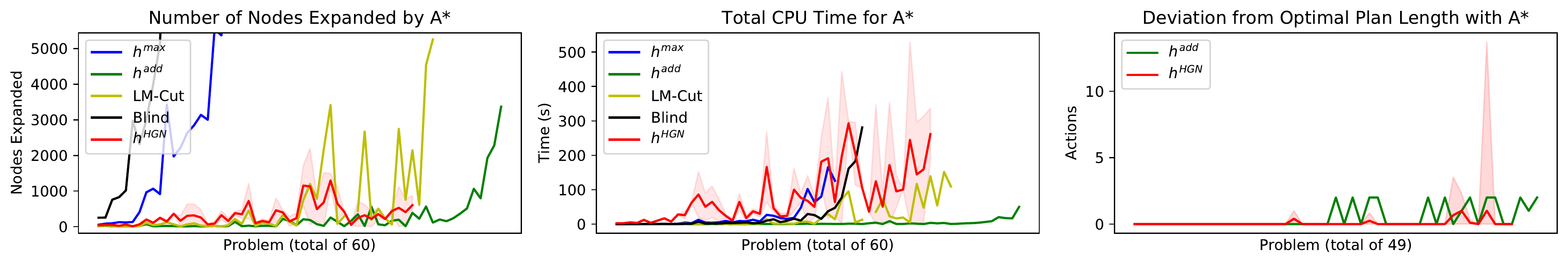}
  \phantomcaption \label{fig:multi-domain-zeno}
\end{subfigure}\\
\plottitle{(g) Domain-independent \textbf{unseen} Blocksworld}
\begin{subfigure}[t]{\textwidth}
  \centering
  \includegraphics[trim={0 0 0 \trimlength},clip,width=\plotlength]{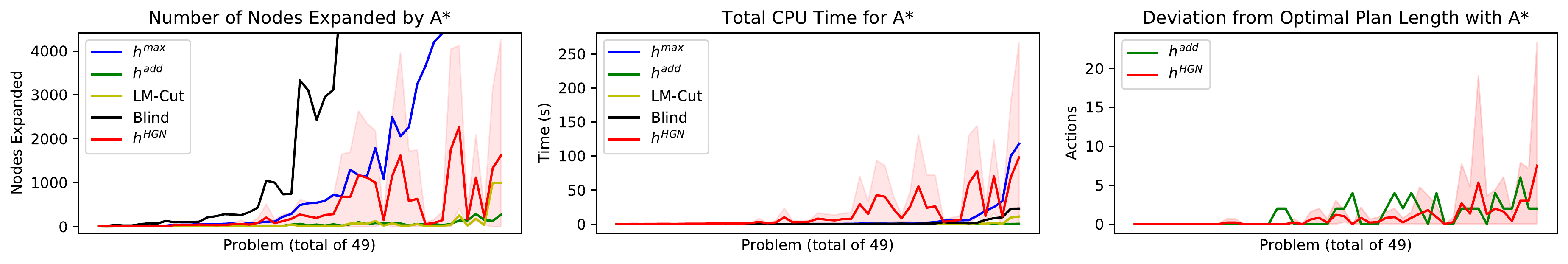}
  \phantomcaption \label{fig:multi-domain-leave-one-out}
\end{subfigure}
\caption{The plots for the results of our experiments described in Section \ref{subsec:results}. The
discontinuities in the curves indicate problems which could not be solved within the search time
limit. The pale red area indicates the 95\% confidence interval for \hSpatial, which may be very
large if our experiments achieved low coverage. Problems for which no optimal solver was able to
find a solution are omitted from deviation from optimal plan length plots.}
\label{fig:main-results}
\end{figure*}

\clearpage